\documentclass[final]{cvpr}

\usepackage{times}
\usepackage{epsfig}
\usepackage{graphicx}
\usepackage{amsmath}
\usepackage{amssymb}
\usepackage{CJK}
\usepackage{xcolor}
\usepackage{subcaption}
\usepackage{url}
\usepackage{threeparttable}
\usepackage{multirow}

\usepackage[pagebackref=true,breaklinks=true,colorlinks,bookmarks=false]{hyperref}

\def\cvprPaperID{9} 
\def\confYear{CV4Animals 2021}

\begin{document}

\title{hSMAL: Detailed Horse Shape and Pose Reconstruction for 
Motion Pattern Recognition}


\author{Ci Li$^1$ ~~~~ Nima Ghorbani$^2$ ~~~~ 
Sofia Broomé$^1$ ~~~~ Maheen Rashid$^3$\\ Michael J.~Black$^2$ ~~~~ Elin Hernlund$^4$ ~~~~ Hedvig Kjellström$^{1,5}$ ~~~~ Silvia Zuffi$^6$\\
$^1$ KTH, Sweden {\tt cil,sbroome,hedvig@kth.se} \\
$^2$ MPI Intelligent Systems, T{\"u}bingen, Germany {\tt nima.ghorbani,black@tuebingen.mpg.de}\\
$^3$ Univrses AB, Sweden {\tt maheen.rashid@univrses.com}\\
$^4$ SLU, Sweden {\tt elin.hernlund@slu.se}~~~~ $^5$ Silo AI, Sweden \\ $^6$ IMATI-CNR, Italy {\tt silvia@mi.imati.cnr.it}}


\maketitle

\begin{abstract}
In this paper we present our preliminary work on model-based behavioral analysis of horse motion. 
Our approach is based on the SMAL model~\cite{zuffi20173d}, a 3D articulated statistical model of animal shape. We define a novel SMAL model for horses based on a new template, skeleton and shape space learned from $37$ horse toys. We test the accuracy of our hSMAL model in reconstructing a horse from 3D mocap data and images. We apply the hSMAL model to the problem of lameness detection from video, where we fit the model to images to recover 3D pose and train an ST-GCN network~\cite{yan2018spatial} on pose data. A comparison with the same network trained on mocap points illustrates the benefit of our 
approach.
\end{abstract}


\vspace{1mm}
\section{Introduction}
\vspace{1mm}

In this paper, we address the task of automatically detecting lameness in horses 
from video. We propose to extract rich 3D information of the horse's shape and pose and perform lameness
analysis on this representation.
Motion information is easy to capture in the form of 2D image keypoints, or, in the case of multi-view calibrated images, as 3D keypoints in space. Here, we go one step further, taking a model-based approach and encoding horse motion as temporal sequences of 3D pose data from the model. We argue that such an animal-centric representation is 
suited for pattern recognition tasks like lameness detection.

\vspace{1mm}

For 3D reconstruction, we employ the SMAL~\cite{zuffi20173d} model, a 3D model for quadruped animals.
SMAL defines a 3D articulated mesh, where the animal shape is generated with a low-dimensional linear model as a vertex deformation from a reference template. The SMAL shape space is learned on a set of quadrupeds of various species. While powerful in the inter-species representation, SMAL is not suited for a detailed intra-species analysis, as the number of training set samples for a specific species is deficient.



In this study, we present a horse-specific version of the SMAL model, which we call hSMAL. The model is based on an accurate basic horse body model and trained with scans of horse toys depicting horses of different breeds and ages.
In order to verify if a 3D statistical shape model learned on toy horses can be an accurate representation for real horses, we investigate the 3D reconstruction accuracy of the hSMAL model using a dataset of horses with simultaneous 3D motion capture and multi-view video.
Our analysis shows that hSMAL can reconstruct the 3D shape of a real animal with an average error of about $7$ cm. In comparison, the average error with the more generic SMAL model is about 14 cm.

We apply the hSMAL model to detect lameness, a pattern recognition task of large relevance to animal welfare. Lameness is defined as ``an irregularity or defect in the function of locomotion'' \cite{liautard1888lameness}, with asymmetry patterns 
in the movement caused by the horse's reflex to reduce the weight on the lame limb. Forelimb lameness can be observed 
through the difference in the head's up-down motion between steps. Hindlimb lameness is more difficult to observe but can be detected through asymmetries in the pelvis motion pattern. 
The lameness detection is performed with the spatial and temporal graph convolution network (ST-GCN)~\cite{yan2018spatial}. We train with sequences of horse motions represented with: hSMAL pose parameters (relative 3D angles of body parts); 3D joint locations of the model; 3D markers positions obtained with a mocap system. The comparison between the network's performance in the different input settings supports our hypothesis that a 3D-model-based representation of 3D pose facilitates the detection of motion patterns.    

\section{Related Work}


\paragraph*{Animal 3D reconstruction.}

Our work builds on the SMAL model~\cite{zuffi20173d}
, which can represent different quadruped species with a unified shape space. 
SMAL has been used by Biggs et al. for dog tracking from video
~\cite{biggs2018creatures} and canine 3D pose and shape estimation~\cite{biggs2020left}, where they add scale parameters to the SMAL model's limbs to perform end-to-end dog reconstruction from monocular images. 
Zuffi et al.~\cite{zuffi2019three} propose an end-to-end reconstruction method also exploiting animal appearance. 
Badger et al.~\cite{badger20203d} focus on capturing the shape and pose of birds, and 
Kearney et al.~\cite{kearney2020rgbd} present a method for dog pose estimation from RGB-D images, introducing a novel dog model with an improved body skeleton where shoulders joints have a translation parameter to adapt to different dog breeds. 

\vspace{-4mm}
\paragraph*{Animal lameness detection.}

Lameness is a pressing welfare problem for horses as it is a 
common disease symptom in the species. 
Accurate and early recognition is important, but has proven to be a challenge also for veterinary experts.
While lameness traditionally is diagnosed via visual subjective judgement by a veterinarian, there has been a recent development of automatic methods using accelerometers \cite{taneja2020machine} and gyroscopes \cite{article,9216873} mounted on the animal. RGB-D imaging has been used for cattle lameness detection \cite{3dVideoCattleDetection}, 
and radar data has also been used \cite{shrestha2017gait} in horse lameness detection.

The method proposed here takes a step further by employing a much richer shape and pose model, thus modeling the animal motion on a more detailed level.

\vspace{-4mm}
\paragraph*{Human gait analysis.}

On the human side, there has been a more extensive development of methods for gait analysis and action recognition. Our lameness detection approach is inspired from the human action recognition work of Yan et al.~\cite{yan2018spatial}, which proposes ST-GCN. ST-GCN can be considered the state-of-the-art approach for human action recognition from skeleton representations.

The method proposed here is the first to perform this type of detailed gait analysis from video in an animal setting.

\section{Method}


\paragraph*{The hSMAL model.}



\begin{figure}[t]
\centerline{\includegraphics[width=\linewidth]{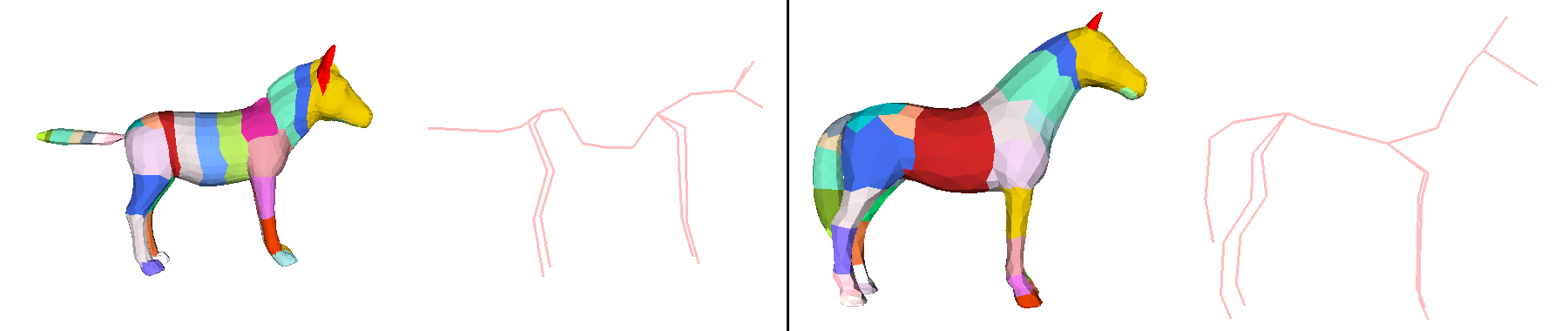}}
\vspace{-2mm}
\caption{From left to right, the SMAL model and its skeleton, the hSMAL model and its skeleton.}
\label{fig:method_SMAL}
\end{figure}

\begin{figure}[t]
\centerline{\includegraphics[width=0.95\linewidth]{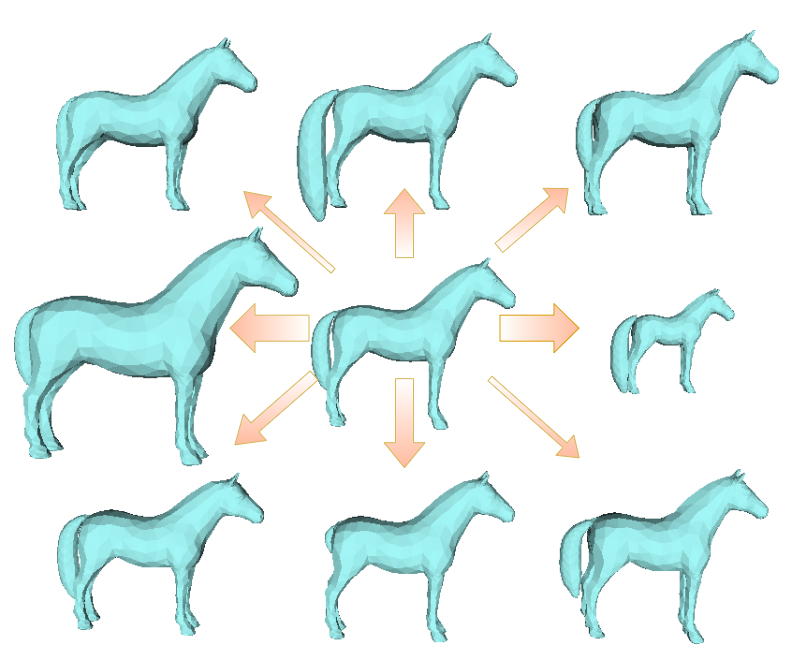}}
\vspace{-2mm}
\caption{The first four principal components in the hSMAL shape space. The arrow width shows from the first to fourth components. All components are shown with $\pm2std$.}
\label{fig:shape_shapce_SMAL}
\end{figure}

The hSMAL model is the horse-specific version of the SMAL model. It defines a 3D mesh as a template deformation, modeling the shape variation, followed by skeleton-based articulation with Linear Blend Skinning (LBS) \cite{loper2015smpl, zuffi20173d}. The model is described as a function $M(\beta, \theta, \gamma)$, where $\beta$ is the shape variable, representing the coefficients of the linear space that defines the vertex-based shape deformations; $\theta$ is the pose parameter, which denotes the relative 3D joint rotations; $\gamma$ is the model translation. In contrast to 
the original SMAL model, we use a new horse template with 1,497 vertices and 36 body segments. We learn the model from a set of 37 horse toys using the procedure described in~\cite{zuffi20173d}. \figref{fig:shape_shapce_SMAL} shows the shape space of the hSMAL model. The first, second and third component mainly capture the body size, the size of tail and the neck of the model, respectively. Moreover, differently from the SMAL model and similarly to~\cite{kearney2020rgbd}, the skeleton joint positions are manually defined to better represent the animal anatomy (\figref{fig:method_SMAL}).


\vspace{-4mm}
\paragraph*{Model fitting.}
In our experiments, we fit the hSMAL model to images, 2D keypoints and 3D motion capture (mocap) data. The fitting to images follows the procedure described in~\cite{zuffi20173d,zuffi2018lions}, which comprises estimating 
the body shape $\beta$, the pose $\theta$, the global translation $\gamma$ given all the data, minimizing 
an energy objective function composed by silhouette and 2D keypoint data terms, and shape and pose regularization terms: 

\vspace{-2mm}
\begin{small}
\begin{equation}
F(\beta, \theta, \gamma) = min \sum_{\lambda \in {kp, sil, \beta, \theta}} \omega_{\lambda} E_{\lambda}(\cdot),
\label{equa:mosh_silhouette}
\end{equation}
\end{small}

\vspace{-2mm}
\noindent where $E_{kp}$ is the 2D keypoint loss, $E_{sil}$ is the silhouette loss, $E_{\beta}$ is the shape prior obtained from the models fitted to the horse toys, and $E_{\theta}$ 
is the pose prior obtained from the models fitted to example animations acquired together with the base horse model. The shape and pose priors are obtained following the same procedure as for the SMAL model \cite{zuffi20173d,zuffi2018lions}.
In our experiments, 2D body keypoints are obtained as projections of 3D mocap markers or 
using DeepLabCut (DLC)~\cite{mathis2018deeplabcut,nath2019using}. DLC keypoints are always used for the tail, for which mocap markers are not available.   
Moreover, we exploit 3D mocap data, adding the mocap loss $E_{mocap}$ to Eq.~(\ref{equa:mosh_silhouette}):

\vspace{-5mm}
\begin{small}
\begin{equation}
E_{mocap}(\beta, \theta, \gamma) = \sum_{j} \rho (\left \| \sum_{i=1 }^{3} \lambda_{j_{i}} V_{j_{i}} - \, ^{C}\textrm{Q}_{j} \right \|_2),
\label{equa:mocap loss}
\end{equation}
\end{small}

\vspace{-2mm}
\noindent where $\rho$ is the Geman-McClure robust error function \cite{geman1987statistical} and $^{C}\textrm{Q}$ is the mocap data expressed in the camera 3D reference system. 
For each mocap marker, we express a manually selected corresponding point on the hSMAL model surface with barycentric coordinates: $\sum_{i=1 }^{3} \lambda_{j_{i}} V_{j_{i}}$. 
For 
multi-view reconstruction, 
silhouette loss terms corresponding to each additional view are added to Eq.~(\ref{equa:mosh_silhouette}).

In previous work, 
SMAL 
is used with single images 
\cite{zuffi20173d,zuffi2018lions}, 
while we here consider videos. Our strategy for model fitting to videos consists of a per-frame fitting without temporal smoothing, with a fixed body shape for the whole sequence estimated on a subset of video frames. 

\vspace{-4mm}
\paragraph*{Lameness detection.}
We now utilize the hSMAL model for a challenging pattern recognition task, lameness detection. We classify a pose sequence as lameness or not using spatio-temporal graph convolution networks (ST-GCN) \cite{yan2018spatial}, which has the benefit, compared to other classification methods, that it efficiently captures the information from the skeleton graph structure in both spatial and temporal dimensions for action recognition. It takes the human skeleton as a graph and connects the same joints on consecutive frames through time. 

\begin{figure}
     \centering
     \begin{subfigure}[b]{0.23\textwidth}
         \centering
         \includegraphics[width=\textwidth]{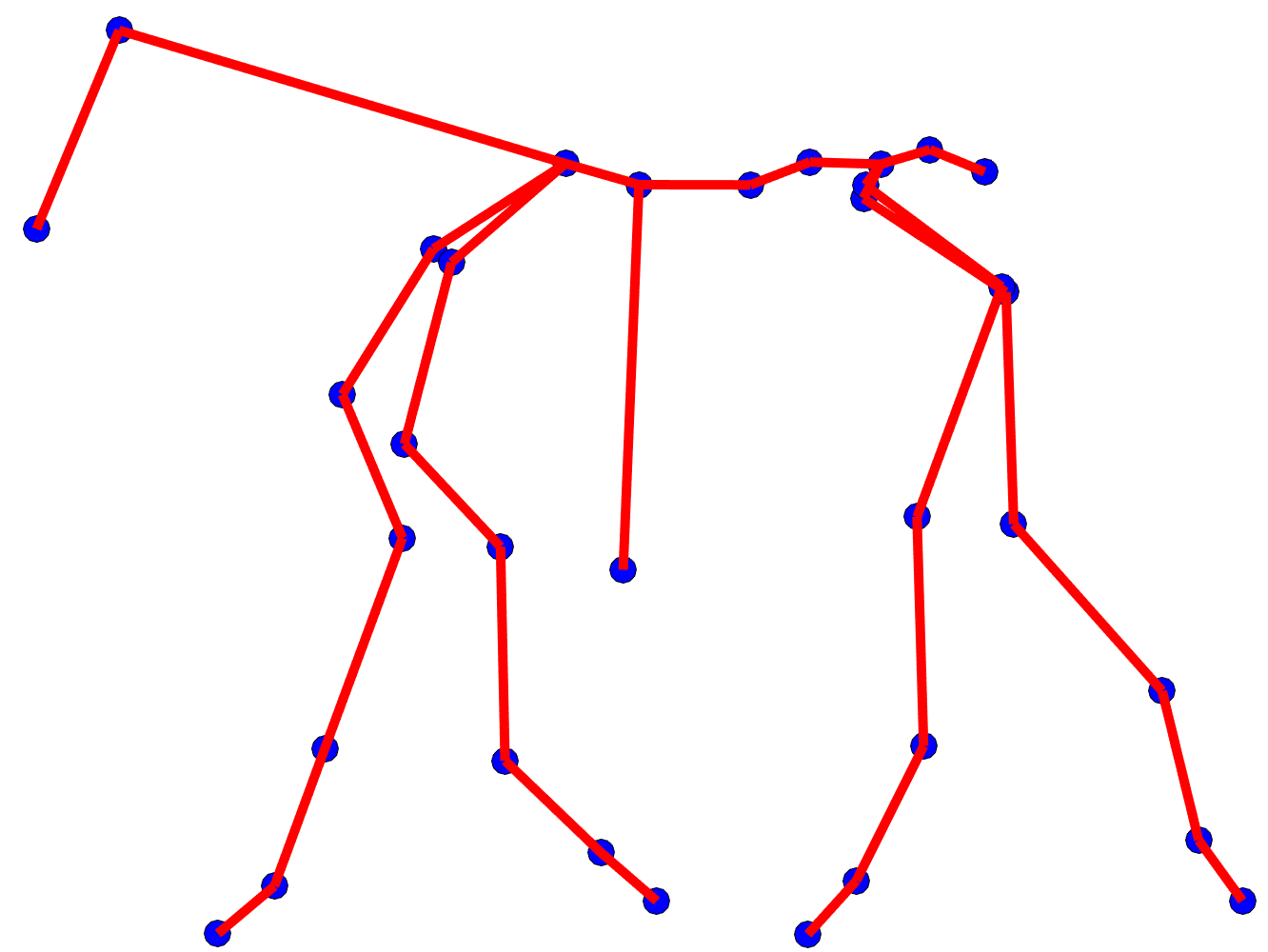}
         \caption{Mocap data.}
         \label{fig:mocap_skeleton}
     \end{subfigure}
     \begin{subfigure}[b]{0.23\textwidth}
         \centering
         \includegraphics[width=\textwidth]{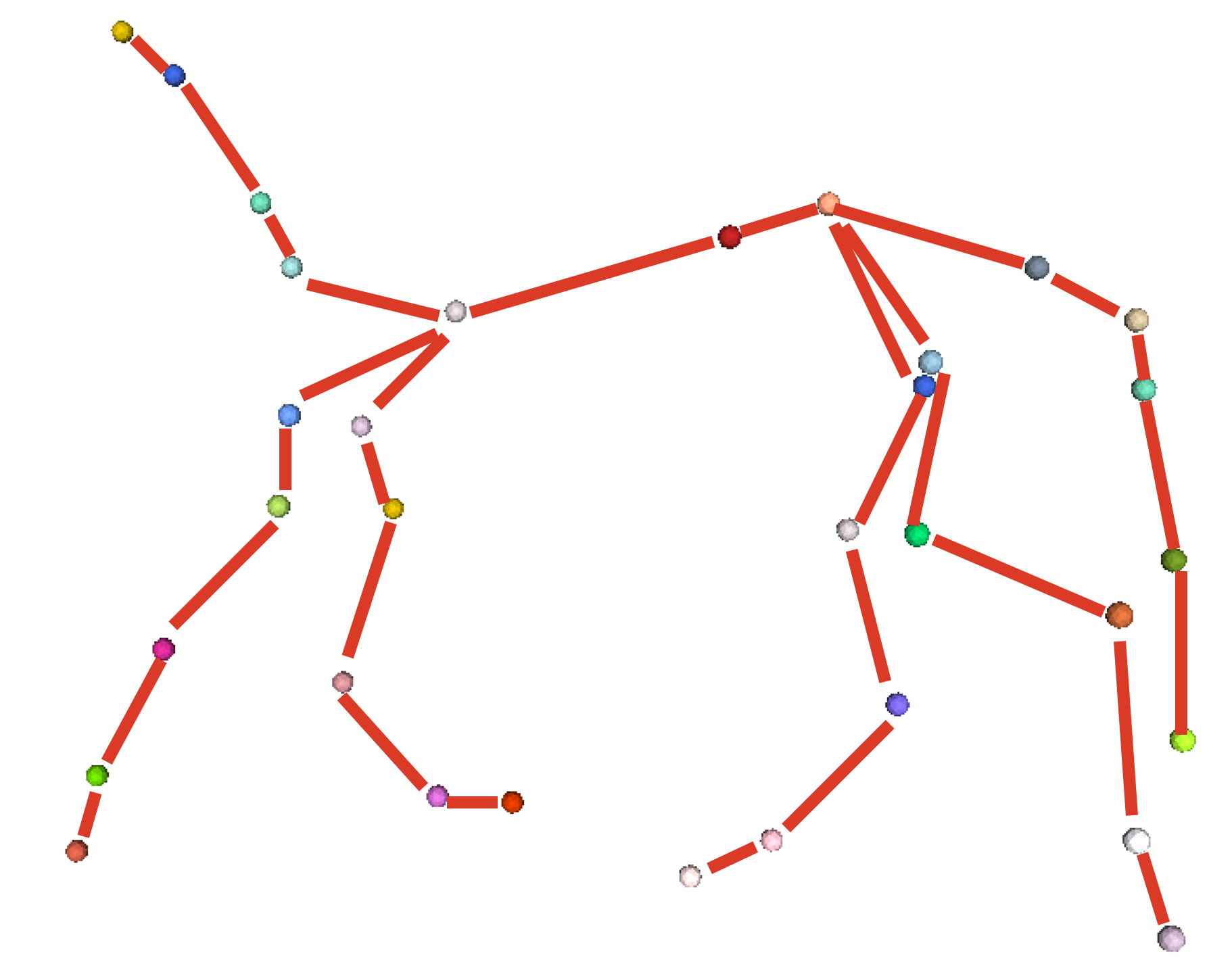}
         \caption{hSMAL model.}
         \label{fig:model_skeleton}
     \end{subfigure}
     
    \caption{The skeleton graph for ST-GCN motion classification.}
    \label{fig:stgcn_graphs}
\end{figure}

To feed the mocap data into ST-GCN, we form the graph in spatial domain \figref{fig:mocap_skeleton}, where the nodes are the mocap markers and the edges are based on the horse structure. For the pose data from the hSMAL model, the graph in the spatial domain is based on the kinematic tree inside the model(\figref{fig:model_skeleton}), where each node is the joint of the model without ears and mouth. The same joints are connected on consecutive frames for both graphs.


The input to the ST-GCN networks are the hSMAL pose parameters, the hSMAL 3D joint coordinates and mocap 3D coordinates, respectively. We utilize the ST-GCN model with four layers of ST-GCN units, and each layer has 64 channels for output. The learning rates are 0.005, 0.1, 0.01, respectively. The random 
feature dropout rate is 0.4 after each ST-GCN unit to avoid overfitting. We train each network for 30 epochs. And we use the model with the highest validation accuracy on the test data. 
The ST-GCN trained with the raw mocap marker positions from the same sequences is the baseline.


\section{Experiments and Results}

\paragraph*{Dataset.}
Our data-driven approach requires access to collected observations of horses with and without lameness. It should be noted that there is a trade-off in collecting as little data as possible and developing as effective methods as possible using that data, which will be of future help for earlier detection and treatment of lameness.

The Horse Treadmill Dataset is collected by the University of Zürich \cite{rhodin2018vertical}.\footnote{Ethical approval for the collection of this dataset (permission number 51/2013) is granted by The Animal Health and Welfare Commission of the canton of Zurich after evaluating the study protocol, and the horse owners gave informed consent for the inclusion of their animals.}
There are video recordings of 10 horse subjects trotting on a treadmill, 
recorded by calibrated and synchronized video from two 
angles and by optical 3D motion capture. Three of the subjects are discarded due to camera calibration problems, leaving seven subjects. 
Each horse is recorded several times without 
and with lameness induced by applying pressure 
to the sole of one hoof per recording (leading to a "stone in the shoe" type of pain). The classification labels are set according to which leg was affected
. This leads to label uncertainty since different individuals experience the pressure differently and may also present with mild baseline motion pattern asymmetries -- a complicating factor for using this classification dataset for method evaluation.


\vspace{-4mm}
\paragraph*{Data preprocessing.}

The One-Shot Video Object Segmentation (OSVOS) framework \cite{caelles2017one} is applied to obtain the silhouettes from each video frame, using five annotated frames per video. 
The mocap data is moreover projected to the side view 
creating accurate 2D marker positions (here called pseudo 2D keypoints). DeepLabCut is used to obtain noisier 2D keypoints for all visible joint locations in the side view, corresponding to a realistic markerless scenario (here called DLC 2D keypoints). The tail is also tracked with DLC, giving a pointset DLC 2D tails. 





\begin{figure}[t]
    \centerline{\begin{subfigure}[b]{0.45\textwidth}
        \includegraphics[width=\textwidth]{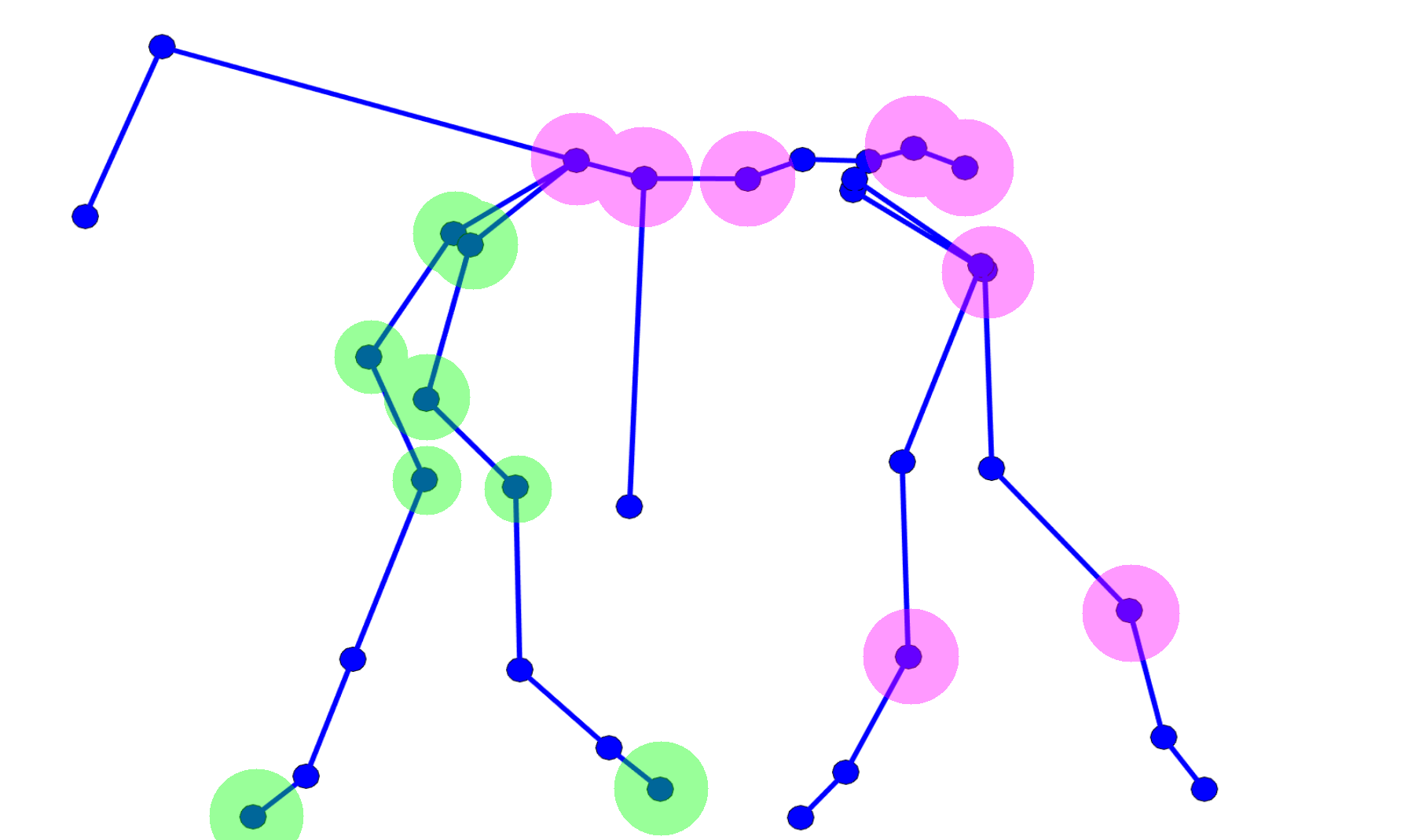}
        \vspace{-6mm}
        \subcaption{SMAL, average error 0.1415 $\pm$ 0.0076 m}
        \label{fig:average_error_SMAL}
	\end{subfigure}}
	\centerline{\begin{subfigure}[b]{0.45\textwidth}
        \includegraphics[width=\textwidth]{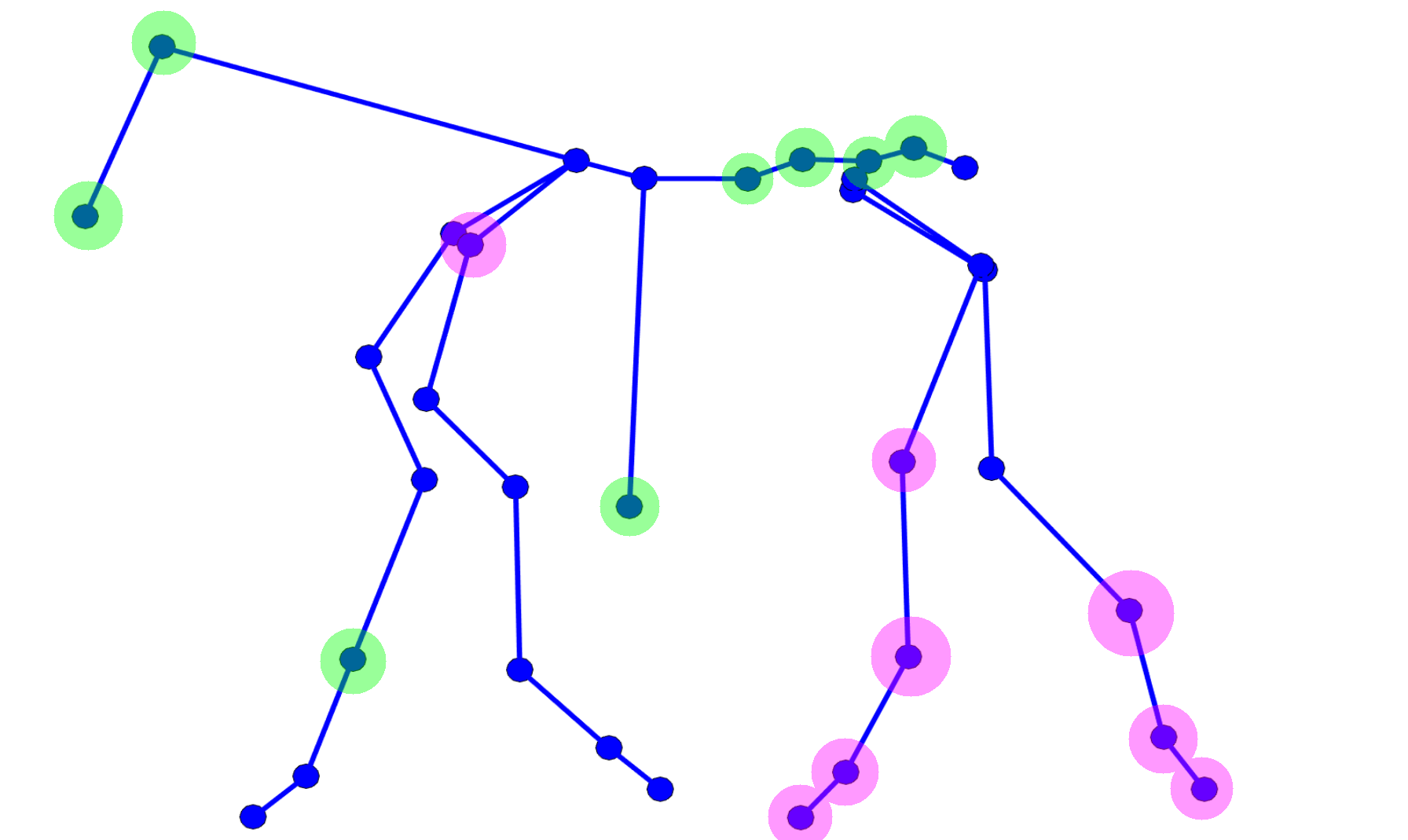}
        \vspace{-6mm}
        \subcaption{hSMAL, average error 0.0696 $\pm$ 0.0021 m}
        \label{fig:average_error_hSMAL}
    \end{subfigure}}
    \caption{Average 3D distance between points and corresponding vertices on the body mesh. Pink = 8 points with largest error, green = 8 mocap points with smallest error.}
    \label{fig:average_error}
\end{figure}

\begin{figure}[!t]
    \centerline{\begin{subfigure}[b]{0.45\textwidth}
        \includegraphics[width=\textwidth]{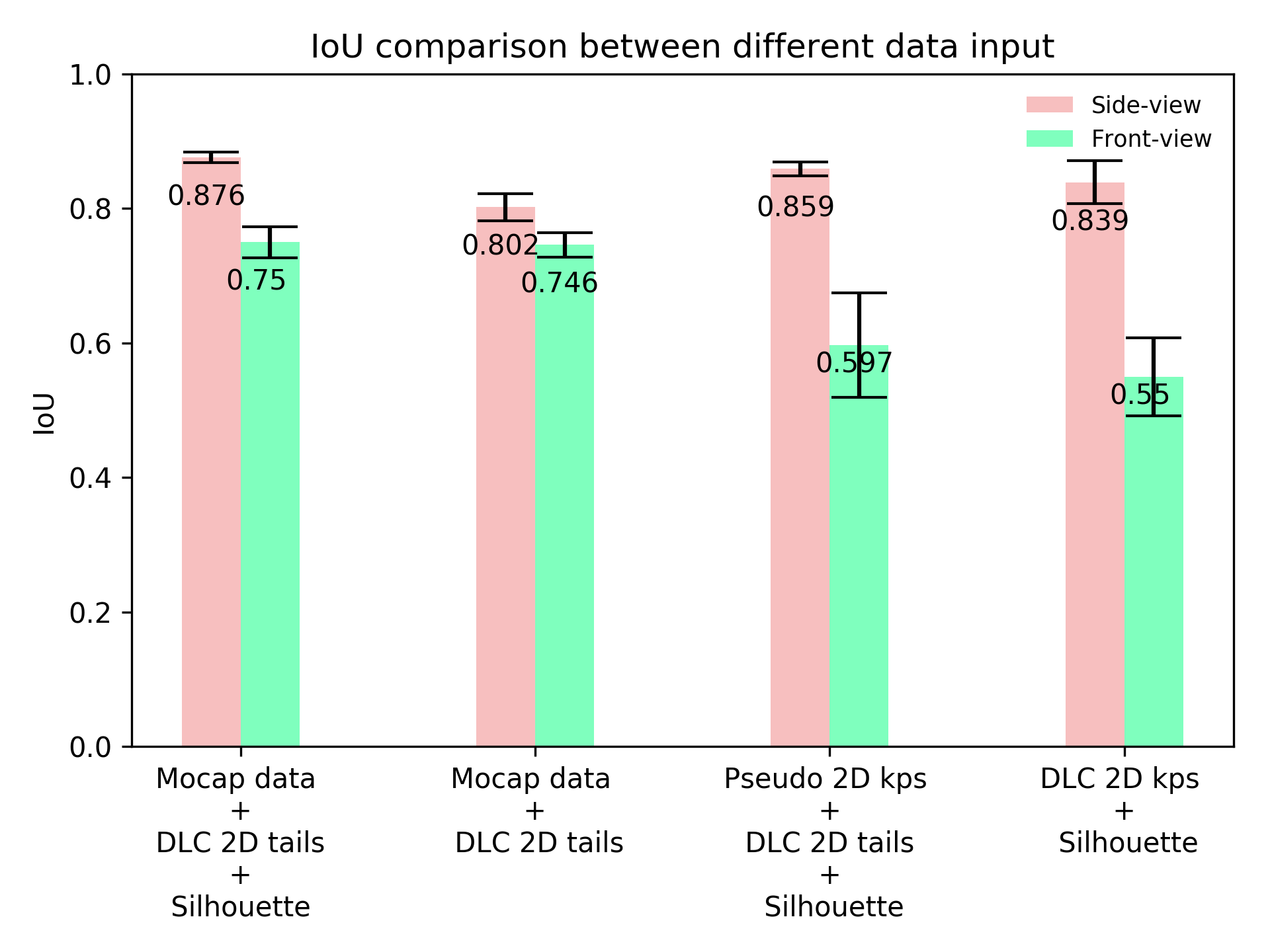}
        \vspace{-7mm}
        \subcaption{Different observations from side-view}
        \label{fig:IoUcompare_different_data}
	\end{subfigure}}
	\centerline{\begin{subfigure}[b]{0.45\textwidth}
         \includegraphics[width=\textwidth]{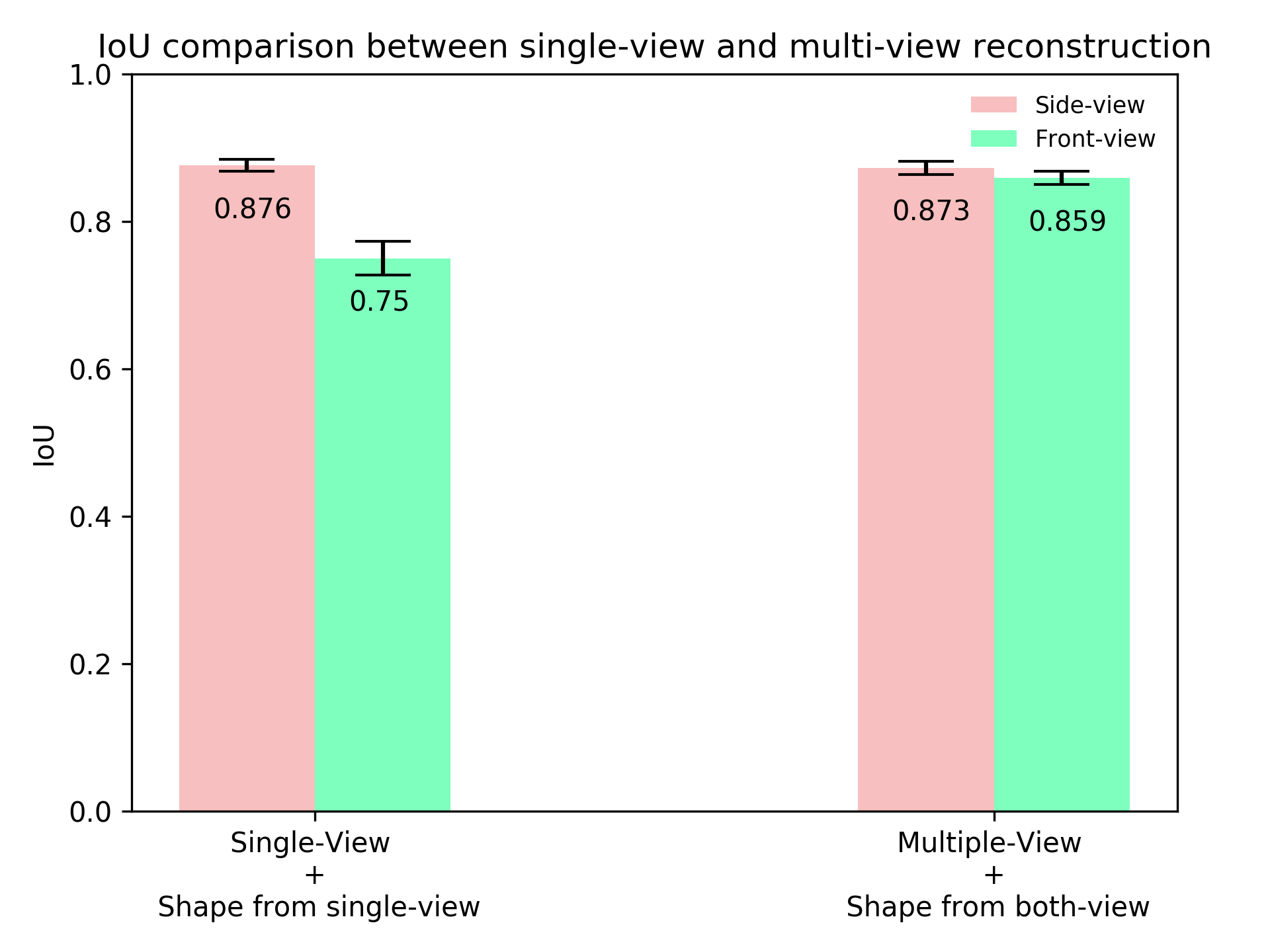}
        \vspace{-7mm}
        \subcaption{Silhouette from side-view and both views.}
        \label{fig:IoUcompare_different_view}
    \end{subfigure}}
    \caption{Model accuracy for different observations. The mean IoU is shown as the bar height and the whiskers show the standard deviation over all frames.}
    \label{fig:IoUcompare}
\end{figure}

\vspace{1mm}
\noindent We evaluate hSMAL at recovering the horse compared to SMAL, and determine the effect of different observations. We then inspect hSMAL to lameness detection. 
\vspace{-4mm}
\paragraph*{Testing hSMAL model accuracy.} 



To test the accuracy of hSMAL in representing a real horse, we choose one recording and select 20 representative frames by using K-means.

We first evaluate hSMAL and SMAL by comparing the average distance between their reconstructed models and the mocap data over the whole recording. The fitting procedure is performed by using 2D keypoints and 3D mocap points, without silhouette adjustment. The shape parameter is first calculated by averaging shape parameters among the 20 selected frames, and kept fixed.
The average (over point and frame index) distance is calculated between the mocap points and their corresponding points on the model. This procedure is repeated for the more generic SMAL model where model fitting is performed using an equine shape prior. 
The results in \figref{fig:average_error} 
indicate that although 
the mocap points on the hind limb are difficult 
for both models to match, hSMAL adapts better to a real horse than SMAL.

\begin{figure*}[t]
\centerline{\includegraphics[width=\linewidth]{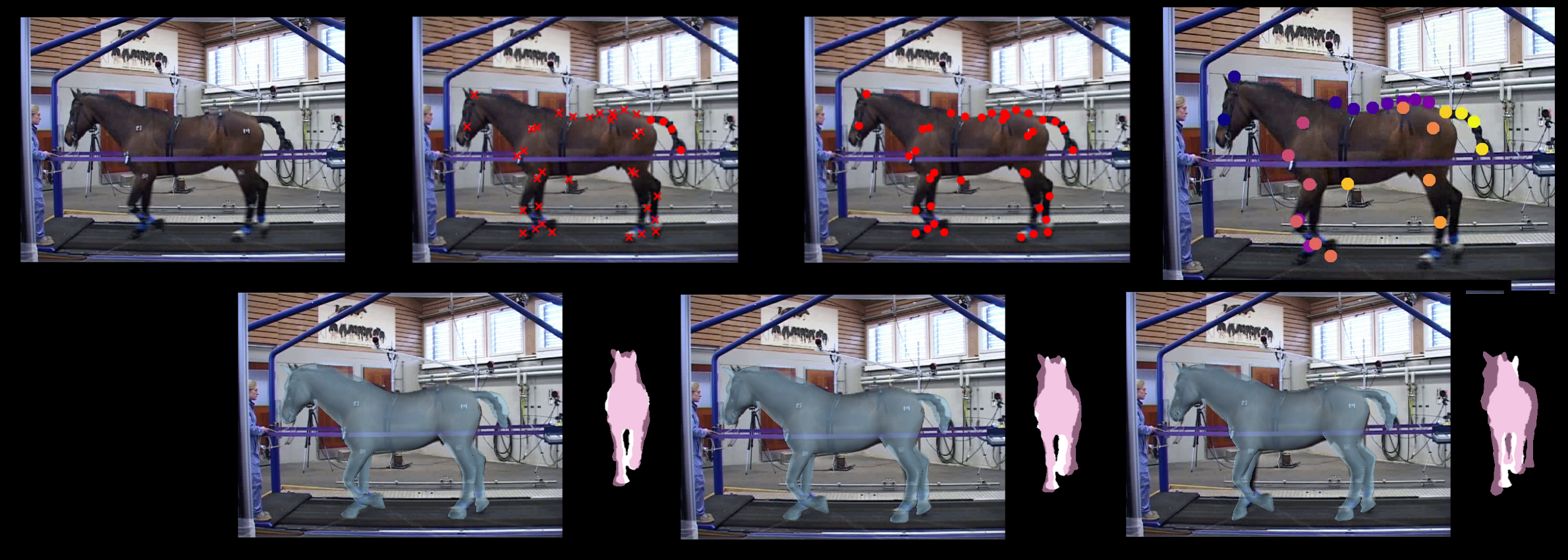}}
\vspace{-1mm}
\caption{Reconstruction results for different inputs. 
Second row from left to right, result for 3D mocap data (blue) and projected silhouette (pink) with the silhouette from OSVOS (white), result for pseudo 2D keypoints and silhouette comparison, result for DLC 2D keypoints and silhouette comparison. DLC 2D tails and side-view silhouette are moreover used in all cases. First row from left to right, video frame, corresponding keypoints in different observation. Video is available on \url{https://vimeo.com/showcase/8402057} and also in the Supplementary Material.}
\label{fig:different2D}
\end{figure*}


We then proceed
to compare the accuracy of hSMAL reconstruction from different inputs.
The reconstruction error is now measured in terms of the Intersection over Union (IOU) 
between the projected silhouette from the reconstructed model and the silhouette from OSVOS in both views. We compare the following input settings: 3D mocap data + DLC 2D tails + side-view silhouette, 3D mocap data + DLC 2D tails, pseudo 2D keypoints + DLC 2D tails + side-view silhouette and DLC 2D keypoints + side-view silhouette. In all the experiments, body shape is fixed to the value estimated from mocap data, 2D keypoints and side-view silhouettes. 
Results over the whole recording are shown in \figref{fig:IoUcompare_different_data}.
As can be expected, the side-view IoUs 
for reconstructions using side-view silhouette 
are higher than those without silhouette. Moreover, the front-view IoU is indeed higher 
when the mocap data is used than when only 2D keypoints are observed. 
Reconstruction examples are shown in \figref{fig:different2D}, along with a link to a video of the entire sequence. 
We also compare the accuracy of 
single-view 
and multi-view reconstruction over the same recording (\figref{fig:IoUcompare_different_view}). The results confirm the expectation that silhouette input improves reconstruction accuracy also for the front-view. 

\vspace{-4mm}
\paragraph*{Lameness detection.}

\begin{table*}[t]
\centering
\caption{Accuracy (\%) of five-class (no, front/hind, left/right) lameness classification for seven subjects. Pose is expressed as mocap point / model pose parameters and joint positions.} 
\vspace{-2mm}
\begin{tabular}{|l|l|l|l|l|l|l|}
\hline
\bf Subject    & \multicolumn{2}{l|}{\textbf{Mocap}}                                                              & \multicolumn{2}{l|}{ \bf \begin{tabular}[c]{@{}l@{}}hSMAL\\ pose parameter($\theta$) \end{tabular}}     & \multicolumn{2}{l|}{ \bf \begin{tabular}[c]{@{}l@{}}hSMAL\\ joint positions\end{tabular}}  \\ \hline \hline
& No augmentation 
& Augmentation 
& No augmentation 
& Augmentation 
& No augmentation 
& Augmentation 
\\ \hline
1       & \textbf{47.8 $\pm$ 9.3 }   & 36.0 $\pm$ 12.0   & 33.7 $\pm$ 6.1   & 35.1 $\pm$ 4.2   & 19.3 $\pm$ 6.2   & 19.2 $\pm$ 3.7 \\ \hline
2       & 38.4 $\pm$ 9.5    & 43.6 $\pm$ 6.8   & 33.9 $\pm$ 6.2   & 32.8 $\pm$ 8.1   & 38.9 $\pm$ 8.9   & \textbf{46.6 $\pm$ 15.8} \\ \hline
3       & 34.7 $\pm$ 13.8   & 28.7 $\pm$ 7.8   & 25.4 $\pm$ 4.5   & 35.4 $\pm$ 4.5   & 29.6 $\pm$ 3.4   & \textbf{40.4 $\pm$ 3.8}\\ \hline
4       & 20.1 $\pm$ 5.5    & 17.1 $\pm$ 9.0  & 23.5 $\pm$ 5.9   & 27.2 $\pm$ 4.2   & 27.5 $\pm$ 7.8   & \textbf{28.0 $\pm$ 5.1}\\ \hline
5       & 24.5 $\pm$ 4.4    & 26.6 $\pm$ 9.4   & 34.0 $\pm$ 10.4  & \textbf{44.1 $\pm$ 6.5}   & 29.4 $\pm$ 6.8   & 31.3 $\pm$ 7.5\\ \hline
6       & 29.9 $\pm$ 8.8    & \textbf{38.8 $\pm$ 6.1}  & 23.6 $\pm$ 4.7   & 22.0 $\pm$ 4.7   & 18.6 $\pm$ 9.3   & 23.9 $\pm$ 7.3\\ \hline
7       & 22.6 $\pm$ 3.9    & 36.4 $\pm$ 11.0  & 31.9 $\pm$ 6.5   & 32.4 $\pm$ 6.1   & 37.1 $\pm$ 8.1   & \textbf{41.5 $\pm$ 6.5}\\ \hline \hline
Average & 31.1 $\pm$ 12.5 &32.5 $\pm$ 12.3 &29.4 $\pm$ 8.0 &32.7 $\pm$ 8.5
&28.6 $\pm$ 10.4 &\textbf{33.0 $\pm$ 12.4 }\\ \hline
\begin{tabular}[c]{@{}l@{}}Average \\ (Without \\Subject1)\end{tabular} & 28.4  $\pm$ 10.7   & 31.9 $\pm$ 12.2  & 28.7 $\pm$ 8.1 & 32.3 $\pm$ 9.0 &  30.2 $\pm$ 10.1 &\textbf{35.3 $\pm$ 11.8} \\ \hline
\end{tabular}
\label{table:result_lameness_w_n_aug}
\end{table*}

We now evaluate the hypothesis that hSMAL provides 
good representation for motion pattern analysis.
This is done by training 
separate ST-GCN classifiers, 
on the sequence of raw 3D mocap point positions, and 
the sequence of pose parameters and joint positions extracted from the fitted hSMAL model from side view reconstruction in each frame. 
Seven experiments are performed according to a cross-validation procedure with 
one subject for validation, one subject for testing, and the rest for training. Data is augmented by mirroring the sequences (creating a left- and right-sided version of every example). We train on the task of five-class classification: 
normal, or lameness in front-left limb, front-right limb, hind-left limb, and hind-right limb, respectively. The random guess accuracy is $20\%$. 

Table~\ref{table:result_lameness_w_n_aug} shows the results with and without data augmentation. Firstly, we can see that data augmentation in general gives a slightly better performance, most probably since this increases the amount of training data.

Secondly, we observe that the hSMAL model, represented either in terms of the pose parameters or joint positions, give the same or slightly better classification performance as the 3D mocap keypoints. This indicates that the hSMAL model fitted to 3D mocap captures all aspects of horse motion encoded in the 3D mocap keypoints, that are of relevance to a challenging pattern classification task such as lameness classification.

Thirdly, although not confirmable from the results, a possible reason for the slight improvement with hSMAL compared to mocap, could be that the hSMAL model not only preserves the information contained in the mocap but even adds information regarding anatomical priors and constraints in the horse body.

\section{Discussion and Conclusion}

This study sets out to capture the detailed shape and pose of real horses through the hSMAL model, and to use the pose data from the model 
to detect motion patterns indicating lameness. 
Our preliminary results indicate 
that the hSMAL model can reconstruct horse shape more accurately than its predecessor, SMAL, and also that the hSMAL pose estimate 
provides a good representation for challenging motion pattern recognition tasks such as lameness detection and classification.

Future work includes optimizing the
marker selection on the mesh 
as in \cite{loper2014mosh} to obtain more accurate locations relative to the mesh. Furthermore, we can improve the current shape space, which is learnt from the toys, with the 3D scan data from 
real horses, 
which will help us establish a more accurate 
hSMAL model. Last, our goal is to apply the model in less constrained 
settings 
without any mocap data, to provide a marker-free and more flexible alternative to the more intrusive wearable sensor approaches that are currently used for motion pattern analysis in horses. 

 
{\small
\bibliographystyle{ieee_fullname}
\bibliography{egbib}
}

\end{document}